\documentclass[11pt]{article}

\usepackage[utf8]{inputenc}
\usepackage{natbib}
\usepackage{mathtools}
\usepackage{color}
\usepackage{graphicx}
\usepackage{hyperref}
\usepackage{float}

\usepackage{multicol, blindtext}

\usepackage{mathtools}
\newtheorem{theorem}{Theorem}
\newtheorem{definition}{Definition}
\usepackage{amssymb, amsmath}
\usepackage{url}

\title{Aligning Intraobserver Agreement by Transitivity}
\author{Jacopo Amidei}
\author{\textbf{Jacopo Amidei}\\
School of Computing and Communications\\ 
The Open University \\
Milton Keynes, UK\\
\texttt{jacopo.amidei@open.ac.uk}}
\date{}

\begin{document}
\maketitle

\begin{abstract}
Annotation reproducibility and accuracy rely on good consistency within annotators. We propose a novel method for measuring within annotator consistency or annotator Intraobserver Agreement (IA). The proposed approach is based on transitivity, a measure that has been thoroughly studied in the context of rational decision-making. The transitivity measure, in contrast with the commonly used test-retest strategy for annotator IA, is less sensitive to the several types of bias introduced by the test-retest strategy. We present a representation theorem to the effect that relative judgement data that meet transitivity can be mapped to a scale (in terms of measurement theory). We also discuss a further application of transitivity as part of data collection design for addressing the problem of the quadratic complexity of data collection of relative judgements.
\end{abstract}

\section{Introduction}

 Annotation reliability plays a pivotal role in data reliability. Krippendorff, in his prominent book \cite{Krippendorff:Krippendorff}, delineates three type of reliability, that is, \textit{stability}, \textit{reproducibility} and \textit{accuracy}. Although IA is strictly speaking a measure of stability, it plays an essential role in reproducibility, which is measured by the Inter-Annotator Agreement (IAA) and accuracy, which is measured calculating the deviations from a given standard. Both reproducibility and accuracy are negatively affected by a low IA. 

The standard method for calculating IA is the test-retest strategy, which is based on the resubmission, after some time, of some items to the annotators. That is, a annotator has to re-assess the same items after some time has elapsed. The comparison of the annotations of the same items provides a measure of the annotator consistency. Test-retest strategy is a measure of the consistency of each of the annotators with themselves over the time, which has the drawback of being time and money consuming. Furthermore, as suggested by \citet[p. 215]{Krippendorff:Krippendorff} test-retest strategy can increase various types of bias which are not strictly related to the annotation task, such as: \textit{carelessness, openness to distractions, or the tendency to relax performance standards when tired}. All of which are amplified by the increase of the annotation time.

In this paper we introduce a new measure of IA based on the concept of \textit{transitivity}. Such a measure can be used in the case of \textit{relative annotations} but not in the case of \textit{absolute annotations}. We recall that, in absolute annotations, human annotators are asked to annotate an item  based on some default ranking or Likert scale. For example, in the case of NLG evaluation, this could involve measuring the grammaticality of a sentence, rating it with a number between 1 to 5. In relative annotations, the human annotators are asked to asses the preference between the subjects under analysis based on some criteria. For example, in the case of NLG evaluation, this could involve choosing between two sentences based on a grammatical preference judgement.  Absolute annotations have the advantage of sup-porting a more fine-grained analysis, for example by using numeric scales,  which is not immediately accessible to relative annotations. However, relative annotations, besides having the convenience of being more intuitive and quicker than absolute annotation \cite{Carterette:Carterette}, have some features which make them very attractive. Firstly, relative annotations allows us to attain higher IAA than absolute annotation does \citep{Jekaterina:Jekaterina, Belz:Belz}. Secondly, in the case of NLG evaluation, absolute annotations, although designed to assess the quality of a system, are often used to compare quality across systems. As we showed in Section \ref{From relative to absolute human judgments}, absolute annotations can be obtained, under some conditions, from relative annotations in a constructive way.\\

\textit{This work is a working in progress, and it presents the theoretical construction of our paradigm. Future developed are presented in the Section} \ref{con}.

\section{Related work}

In this paper, we suggest considering a new kind of annotator consistency in the case of annotations based on preference choice, as for instance in relative judgments evaluations.  We propose to use the property of \textit{transitivity} as a measure of annotation stability. To the best of our knowledge this is an original contribution. However, several papers inspired our proposal. \\

The idea of transitivity as a measure of consistency is not new and can be found for example in \cite{siegelCastellannonparametric}. 

The classical concept of rationality as defined in  Decision Theory (see for example \cite{Luce:Luce}), use the property of transitivity as basic for the concept of rational preferences. Following classical Decision Theory, rational preferences have to be transitive.   Due to the fact that transitivity is ordinal in nature, our proposal is introduced with regard to relative annotation methodologies. Some advantages of using relative annotations instead of absolute annotations are presented, for example, by \citet{Carterette:Carterette}, \citet{Belz:Belz} and \citet{Jekaterina:Jekaterina}. As shown in \citet{Carterette:Carterette} and \citet{Belz:Belz}, such methodologies are more intuitive and quicker than absolute annotations. Furthermore, as shown in \citet{Jekaterina:Jekaterina} and \citet{Belz:Belz}, relative annotations  reach a higher IAA than absolute annotations. Recently preferences annotations have been investigated as an alternative to absolute annotations in the areas of machine translation and information retrieval systems: see for examples, \citet{vilar2007human}, \citet{Rorvig:Rorvig},  \citet{Carterette:Carterette}, \citet{song2011select} and \citet{bashir2013document}. 

Finally, taking inspiration from Measurement theory \citep{Roberts:Roberts} we show how to infer absolute annotations from relative annotations. A similar result was introduced by \citet{Rorvig:Rorvig} using the concept of Simple Scalability. Whereas the Simple Scalability uses the property of transitivity, substitutibility and independence, the representation theorem we present uses transitivity and completeness. 

\section{Using transitivity to measure IA}\label{Using transitivity to measure intraobserver agreement}

The test-retest strategy is aimed at determining what we can refer to as logical consistency. If an annotator prefers subject $i_1$ over subject $i_2$ then (s)he is not consistent if at the same time (s)he prefer subject $i_2$ over subject $i_1$.  From a more general point of view, we can state that an annotator is inconsistent if his(her) claims are not compatible with each other, where this incompatibility can be also of a different nature than the logical one.
For example, given the subjects $i_1$, $i_2$ and $i_3$, the following claims are incompatible between them:  
\begin{enumerate}
    \item $i_1$ is  preferred to $i_2$;
    \item $i_2$ is preferred to $i_3$;
    \item $i_3$ is preferred to $i_1$.
\end{enumerate}
The property in question is known as \textit{transitivity}. Given the subjects $i_1$, $i_2$ and $i_3$, transitivity states that if $i_1$ is  preferred to $i_2$, and $i_2$ is preferred to $i_3$, then $i_1$ is preferred to $i_3$. In classical decision theory (see for example, \cite{Luce:Luce}), transitivity plays a pivotal role in defining the concept of rational preference under the \textit{normative} interpretation. Indeed, following this interpretation, a rational man should make judgments that are transitive. Although transitivity has raised several discussions about its adequacy as a property of rationality, see for example \cite{Fishburn82:Fishburn82} and  \cite{Regenwetter:Regenwetter}, we believe that it can be safely used as a measure of the annotators' internal consistency.

It is important to say that, in the annotation tasks that we are interested in this paper, annotators have to assess items based on the same criteria. For instance, suppose that the annotation consists of relative annotations about the sentences  $s_1$, $s_2$ and $s_3$.  It could happen that an annotator prefers sentence $s_1$ to sentence $s_2$ and sentence $s_2$ to sentence $s_3$, based on the sentences' grammaticality, and at the same time sentence $s_3$ to $s_1$, based on the sentences' fluency. However, in this case, the annotator is assessing the quality of the three sentences based on different criteria, which compromises the object of the specific evaluation. For example, evaluating which system generated grammatically better sentences. Note that because our use of transitivity assumes that preferences are made using constant fixed criteria, where the annotation criteria are very general (e.g. overall quality) and prone to unstable interpretation, our measure is less appropriate. 

As the above example highlights, the property of transitivity can be used to assess the IA. Indeed, let us suppose that we have a not-transitive preference in an evaluation about grammaticality. So, in this case, an annotator declares that $s_1$ is more grammatical than $s_2$, $s_2$ is more grammatical than $s_3$ but $s_3$ is more grammatical than $s_1$. In this case we cannot reach consistent conclusions about the grammaticality of the sentences $s_1$, $s_2$ and $s_3$. If there are many inconsistencies in the annotator's preferences, this weakens the basis for comparing systems based on those preferences.

Differently from the test-retest strategy, transitivity is part of a single test/annotation scenario, which reduces the cost of the annotation. It can also reduce subjective bias linked to the time elapsed, for instance, the tendency to relax performance standards when tired. This is especially true in a procedure that resubmits the items during the same annotation.\footnote{In a preference annotation task we are interested in, an item is made by a couple of subjects from which annotators have to express a preference.} Such a procedure made use of the annotator's time to judge again items already annotated. When using the transitivity we can directly measure the IA by taking triplets of subjects for which annotators have given pairwise preferences during the annotation. In this case we can save the time and the money usually involved in the second annotation.
 
Additionally, the property of transitivity can play an important role in the area of evaluation of NLG systems. As we show in the Section \ref{From relative to absolute human judgments}, transitivity can be used, alongside the property of completeness\footnote{This property ensures that, given two subjects $i_1$ and $i_2$, if $i_1$ is different from $i_2$ then the annotator has to prefer  $i_1$ over $i_2$ or $i_2$ over $i_1$ or can express indifference preference between $i_1$ or $i_2$. Given the finiteness of the set of sentences to be assessed, although the total number of the sentences to be split between the annotators can be quite large, the completeness seems a reasonable requirement for annotation tasks.}, to obtain absolute annotations from relative annotations.

\subsection{How to calculate the IA with transitivity}\label{How to calculate the IA with transitivity}

Preference annotation can be \textit{strict} or \textit{weak}. In strict preference annotation, given two subjects $i_1$ and $i_2$, the annotator is requested to express one and only one of the following preference: either $i_1$ is preferred to $i_2$ or $i_2$ is preferred to $i_1$. In the case of weak preference the two alternatives can be chosen together, that is, $i_1$ is preferred to $i_2$ and $i_2$ is preferred to $i_1$. In this case an annotator expresses equal preference between the two subjects. In this section we consider the case of weak preference, which gives more freedom to the annotators. Indeed, in the case of strict preference, the annotators are forced to give a preference, as well as a case when they do not have a clear preference between two subjects. 
In order to measure the transitivity, all the judgements are performed within pairwise preferences of triplets of subjects $i_1$, $i_2$ and $i_3$. That is, ($i_1$, $i_2$), ($i_1$, $i_3$) and ($i_2$, $i_3$).  

The standard methods used for measuring IAA and IA can be used in our case. Since the pivotal work of \citet{Carletta:Carletta}, kappa coefficients $K$ are used for measuring annotation agreement. Using the more general formulation, as given by \citet{Carletta:Carletta}, the kappa coefficient $K$ can be expressed with the following formulation:
$$K = \dfrac{P(A) - P(E)}{1 - P(E)}$$
where $P(A)$ is the proportion of times the annotators agree, whereas $P(E)$ is the proportion of times the annotators would be expected to agree by chance. In our case, the IA is calculated for each annotator based on each triplet of items between which (s)he have to express his(her) preferences. Consequently, $P(A)$ is the proportion of times that the annotator is transitive, and $P(E)$ is the proportion of times that s(he) would be transitive by chance, that is $0.48$. Such a number is calculated by counting the proportion of times the annotator is transitive by chance versus not transitive by chance. In a preference annotation task, given two subjects $i_1$ and $i_2$, an annotator can express three preferences:  $i_1$ is preferred to $i_2$ (in symbol $i_1 <  i_2$), $i_2$ is preferred to $i_1$ (in symbol $i_2 <  i_1$) or $i_1$ and $i_2$ are equally preferred (in symbol $i_2$ $\backsim$  $i_1$). For the sake of simplicity, let us merge the symbols $<$ and $\backsim$  in the symbol $\lesssim$. Given three subjects $i_1$, $i_2$ and $i_3$ there are eight possible preference judgements.
\begin{enumerate}
\item $i_1 \lesssim i_2$ and $i_2 \lesssim i_3$ and $i_1 \lesssim i_3$.
\item $i_1 \lesssim i_2$ and $i_3 \lesssim i_2$ and $i_1 \lesssim i_3$.
\item $i_1 \lesssim i_2$ and $i_2 \lesssim i_3$ and $i_3 \lesssim i_1$.
\item $i_1 \lesssim i_2$ and $i_3 \lesssim i_2$ and $i_3 \lesssim i_1$.
\item $i_2 \lesssim i_1$ and $i_2 \lesssim i_3$ and $i_1 \lesssim i_3$.
\item $i_2 \lesssim i_1$ and $i_3 \lesssim i_2$ and $i_1 \lesssim i_3$.
\item $i_2 \lesssim i_1$ and $i_2 \lesssim i_3$ and $i_3 \lesssim i_1$.
\item $i_2 \lesssim i_1$ and $i_3 \lesssim i_2$ and $i_3 \lesssim i_1$.
\end{enumerate}
We note that for each preference above the symbol $\lesssim$ have two meaning. Indeed, it can be interpreted either as $<$ or $\backsim$. Consequently, each of the eight points above can be split into 8 different combinations of $<$ or $\backsim$. This means that we have 64 possible combinations at the end. Over these we get 37 repetitions. 
For example, from both the point 1 and 2 we get the preferences:
$i_1 < i_2$ and $i_2 \backsim i_3$ and $i_1 < i_3$. There are 13 transitive assessments left out of 27 possibilities. That is, $13/27 =  0.48$ is the probability that an annotator is transitive by chance. 

We note that in the case of strict preference the value of $P(E)$ has to be considered $0.75$. Indeed, by using the same methodology we can see that there is a total of 6 transitive assessments out of 8. Indeed, if we only use the symbol $<$ we can see that only points 3 and 6 are not transitive. This fact makes our measure not suitable for the case of strict preferences. We note that this can explain the conclusion reached by \cite{hui2017transitivity}, where strict preference is considered transitive across annotators, whereas weak preference is considered not transitive.

Let us give an example of how to calculate the IA.  We remind that, in order to measure the transitivity, all the judgements are performed within pairwise preferences of triplets of subjects $i_1$, $i_2$ and $i_3$.  Suppose the annotation is performed by three annotators $A_1$, $A_2$ and $A_3$ on the triplets of items $(i_1, i_2, i_3)$, $(i_4, i_5, i_6)$ and $(i_7, i_8, i_9)$. This mean that annotators have to give preference between the pairwise of subjects in the triplets  $(i_1, i_2, i_3)$, $(i_4, i_5, i_6)$ and $(i_7, i_8, i_9)$. Table \ref{Annotators' preference assement. T means transitive preferences and NT means not transitive preferences.} reports the artificial annotations.
\begin{table}[h!]
\center
\begin{tabular}{ |p{1.7cm}|p{1.5cm}|p{1.5cm}|p{1.5cm}|}
 \hline
Annotator &  $(i_1, i_2, i_3)$ & $(i_4, i_5, i_6)$ & $(i_7, i_8, i_9)$\\
 \hline
 \hline
$A_1$ & T & T & T\\
 \hline
$A_2$ & T & NT & T\\
 \hline
$A_3$ & NT & T & NT\\
 \hline
\end{tabular}
\caption{Annotators' preference assessment. T means transitive preferences and NT means non transitive preferences.}
\label{Annotators' preference assement. T means transitive preferences and NT means not transitive preferences.}
\end{table}
The annotators' IA are reported in the last column of  Table \ref{Annotator's IA}.
\begin{table}[h!]
\center
\begin{tabular}{ |p{1.7cm}|p{1cm}|p{1cm}|p{1cm}|}
 \hline
Annotator &  $P(A)$ & $P(E)$ & $K$\\
 \hline
 \hline
$A_1$ & 3/3 & 0.48 & 1\\
 \hline
$A_2$ & 2/3 & 0.48 & 0.34\\
 \hline
$A_3$ & 1/3 & 0.48 & -0.28\\
 \hline
\end{tabular}
\caption{Annotators' IA. We remind that $K$ values range from 1 (perfect agreement) to -1.}
\label{Annotator's IA}
\end{table}

\section{The use of transitivity in experimental design}

The transitivity property can also be considered from a normative point of view, as in Decision Theory. This allows thinking about it as a condition to guarantee a specific idea of annotators' consistency as stability. In this case transitivity is not used to check the annotators' internal consistency, but rather it is assumed. Of course, we may only want to make such an assumption based on evidence. For instance, we can first test annotators' consistency on a sample of our dataset. Once we have established consistency on the sample (or eliminated inconsistent annotators), we can then work with the assumption of consistency for the remaining data -- this practice is also common when it comes to the application of IAA.

If we assume consistency, we can drastically reduce the number of annotations we request from annotators. As an example, we can imagine an interactive software which removes the couple, whose order can be deducted from the assessment done before, from the set of couples. For instance, coming back to the example with the three sentences $s_1, s_2$ and $s_3$, let's suppose that an annotator prefers sentence $s_1$ over sentence $s_2$ and sentence $s_2$ over sentence $s_3$. Then, in this case, the software does not present the couple of sentences $s_1$ and $s_3$ and infers from the annotation before that the annotator prefers sentence $s_1$ over sentence $s_3$. This annotation task design, besides guaranteeing the transitivity of the evaluation, allows us to reduce the problem of the quadratically explosion of the possible alternatives, which is present in the relative annotation evaluation \citep{Carterette:Carterette}. This has the advantage of reducing the time required to complete the annotation. 

\section{From relative to absolute human annotations}\label{From relative to absolute human judgments}

Inspired by Measurement Theory \citep{Roberts:Roberts}, in this section, we show how to extract absolute annotations from relative annotations. 

Let us begin by giving a definition. 

\begin{definition} \label{definition tran com}
Let $A$ be a set. A \textit{binary relation} $R$ on $A$ is:
\begin{itemize}
\item[] \textit{Strongly complete} if for each $a, b \in A$ such that $a \neq b$, $aRb$ or $bRa$;
\item[] \textit{Transitive} if $aRb$ and $bRc$ then $aRc$, for each $a, b, c \in A$. 
\end{itemize}
\end{definition}
An example of strongly complete relation is the strict order ($<$) between natural numbers. 

It is commonly believed that it is possible to measure some entity if we can associate, in a default fashion, a number to it. More generally, in order to measure a set of entities, we would like to have a function which associated with each of them a real number. For example, given a set of people, suppose we are interested in measuring their weight. In this case we would like to have a tool, for instance a weight scale, which associates a number with each person of the set in a compact manner. Such number will be interpreted as the weight of that person. Suppose now we are interested in measuring the weight of those people, instead of associating a number with them, by ordering them from the lightest to the heaviest. Both of the approaches are informative about the weight of the people. The questions that arises are: Are these two approaches correlated? How do they correlate? A goal of Measurement Theory is to explain under which conditions it is possible to answer these questions. The general idea is to find properties (technically called axioms) satisfied by the order, such that it is possible to define a numerical function which mirrors the order. The proof that sanctions this result is called a \textit{representation theorem}. Theorem \ref{theorem} is an example of representation theorem. 

\begin{theorem}\label{theorem}
Let $A$ be a set and $R$ a binary relation on $A$. There exist a real function $f$ on $A$ which, for each $a, b \in A$, satisfies the following:
\begin{center}
$aRb$ if and only if $f(a) \geq f(b)$
\end{center}
if and only if $R$ is transitive and strongly complete.
\end{theorem}
Although we do not report the proof, which can be found in \cite[p. 107]{Roberts:Roberts}, we present the construction of $f$. The function $f$ is defined as follows:
\begin{center}
$f(x)$ = the number of $y$ in $A$ such that $xRy$.
\end{center}
In words, $f$ counts how many times an element in $A$ is preferred to some other element in $A$ which is related by $R$.  An example can illustrate this definition. Let $A$ be the set $\{a, b , c\}$ and let $R$ be the set $\{(a, b), (a, c), (b, c)\}$ then $f$ is defined by:
\begin{itemize}
\item[] $f(a) = 2$; $(a, b), (a, c)$
\item[] $f(b) = 1$; $(b, c)$
\item[] $f(c) = 0.$
\end{itemize}

The function $f$ can be interpreted as absolute annotation. We note that any linear transformations of $f$ satisfy Theorem \ref{theorem}. For example, each natural number $n$ different from 0, $f\cdot n$ satisfies Theorem \ref{theorem}. Given its construction, we suggest using the $f$ as presented above to be considered as absolute annotation derived from a relative one.

\section{Limits of the present proposal}

We have shown some advantages of the property of transitivity as a measure of IA. Let us now present a couple of the limits of our proposal. 

On one hand, the use of transitivity as a measure of IA can be used only in the case of relative annotations, whereas it cannot be used for absolute annotations. Although several papers  (see for example \cite{Jekaterina:Jekaterina},  \cite{Belz:Belz} and  \cite{Carterette:Carterette}) demonstrate advantages of relative over absolute annotations, the latter are used much more widely. Additionally, as we observed in Section \ref{Using transitivity to measure intraobserver agreement}, one limit is linked to the fuzziness of the criterion annotated. As with the sorites paradox (or the paradox of the heap), this problem arises from vague predicates. If the criterion is too general then transitivity can fail due to the differences in the aspect used by the annotators to assess different items. This cannot be strictly considered as a failure of annotator internal consistency as stability.

\section{Conclusion and future work}\label{con}

In this paper we have introduced a new approach for checking the annotator IA. Inspired by the concept of rational preference as defined in classical decision theory, we suggested using the property of transitivity to check the annotators' stability. 
We presented some advantages introduced by the concept of transitivity with respect to the test-retest strategy, among which the possibility of using some results from Measurement Theory to constructively derive absolute annotations from relative annotations is present.  
Furthermore, from a normative point of view, assuming transitivity can allow for more efficient collection of human annotations (avoiding the quadratic explosion of annotations that would otherwise be required) as observed originally by \cite{Carterette:Carterette} in the context of relevance annotations for information retrieval. \\

This paper presents the theoretical construction of our paradigm. Future developments consist of:
\begin{enumerate}
    \item  An extensive study and evaluation of our theoretical ideas;
    \item the use of possible weight in the preference annotation, representing the intensity of annotators' preference. 
\end{enumerate}

\bibliographystyle{apalike}
\bibliography{references}

\end{document}